Scientific
Research
Publishing

# MRF-Based Multispectral Image Fusion Using an Adaptive Approach Based on Edge-Guided Interpolation

**Mohammad Reza Khosravi[1,2]\*, Mohammad Sharif-Yazd[3], Mohammad Kazem Moghimi[4], Ahmad Keshavarz[2], Habib Rostami[5], Suleiman Mansouri[2]**

[1]Department of Electrical and Electronic Engineering, Shiraz University of Technology, Shiraz, Iran
[2]Department of Electrical Engineering, Persian Gulf University, Bushehr, Iran
[3]Department of Electrical Engineering, Yazd Branch, Islamic Azad University, Yazd, Iran
[4]Department of Electrical Engineering, Najafabad Branch, Islamic Azad University, Najafabad, Iran
[5]Departmentof Computer Engineering, Persian Gulf University, Bushehr, Iran
Email: *m.khosravi@sutech.ac.ir, *m.khosravi@mehr.pgu.ac.ir

## Abstract

In interpretation of remote sensing images, it is possible that some images which are supplied by different sensors become incomprehensible. For better visual perception of these images, it is essential to operate series of pre-processing and elementary corrections and then operate a series of main processing steps for more precise analysis on the images. There are several approaches for processing which are depended on the type of remote sensing images. The discussed approach in this article, *i.e.* image fusion, is the use of natural colors of an optical image for adding color to a grayscale satellite image which gives us the ability for better observation of the HR image of OLI sensor of Landsat-8. This process with emphasis on details of fusion technique has previously been performed; however, we are going to apply the concept of the interpolation process. In fact, we see many important software tools such as ENVI and ERDAS as the most famous remote sensing image processing tools have only classical interpolation techniques (such as bi-linear (BL) and bi-cubic/cubic convolution (CC)). Therefore, ENVI- and ERDAS-based researches in image fusion area and even other fusion researches often don't use new and better interpolators and are mainly concentrated on the fusion algorithm's details for achieving a better quality, so we only focus on the interpolation impact on fusion quality in Landsat-8 multispectral images. The important feature of this approach is to use a statistical, adaptive, and edge-guided interpolation method for improving the color quality in the images in practice. Numerical simulations show selecting the suitable interpolation techniques in MRF-based images creates better quality than the classical interpolators.

## Keywords

## 1. Introduction

Processing of remote sensing images is a way for achieving information in different usages of geosciences. These images are widely used in different fields such as physical geography and satellite photogrammetric studies, study of climate change, earth physics and earth quick engineering, hydrology and soil erosion, jungle studies and different fields of agriculture. Remote sensing images contain different types such as ground photos, airborne photos and satellite images. Satellite images can be divided into the images in the visible region (optical images), thermal images (infrared), radar images and laser images and so on. For better visual perception of the images, it is essential to operate series of pre-processing and elementary corrections and then operate a series of main processes for more precise analysis on the image [1]. Each of these images is provided with a special means and used for some special usages. For example visible images are appropriate in many applications which need to the natural color. The problem of visible images is that all the surface information is not reachable and this is a reason for using other sensors' images. In visible area, image quality is inappropriate in bad climate and dark night, because passive imaging systems are used in the most optical sensors. Thermal images are often provided in infrared (IR) frequency spectrum and are usually used in warming studies and earth temperature changing. The major problem of these images is the lower spatial accuracy of IR sensors in comparison to the optical sensing. However, radar images which are often obtained from active radars have wide usages in the applications which the optical sensing is not suitable, e.g. extracting objects, buildings and roads, exploration if natural resources and identifying facilities and complication of earth surface [2].

Several remote sensing satellites have been launched for different missions. One or more sensors are placed on each of these satellites for different purposes. Regarding the usages, each sensor takes images in some special frequency bands and gives us the proportional information. Landsat satellite series has been made from a group of 8 satellites (presently) by incorporation of NASA and USGS and they have been launched since 1972. Recently just Landsat-7 and Landsat-8 satellites are active (available) which are launched in 1999 and 2013, respectively.

The most popular satellite of this group is Landsat-7 and its multispectral sensor that named ETM+ is one the most popular remote sensing sensors in the world. Landsat-8 satellite has been considered for at least a ten-year mission and two multispectral sensors were placed on it. These two sensors were named OLI and TIRS and they provide images in nine and two frequency bands, respectively. The OLI sensor provides multispectral images which contain almost all of the ETM+ bands. However, they have been improved in SNR and spatial resolution. This sensor takes images in visible and IR, and the TIRS sensor which is considered as a thermal sensor, just takes images in two IR bands. Some of the OLI sensor's bands are presented in Table 1. Lansat-8 collects spatial data with medium resolution (30 meters). The TIRS thermal sensor has a 100 meters resolution in both of its frequency bands where this resolution is considered a weak accuracy. Landsat-8 images are freely available and their format is Geo-TIFF, a file format for lossless storage which is geo-referenced. Therefore, they are not required to the geometrical corrections. Combination of 2nd, 3rd and 4th bands makes a visible image with 30 meters resolution and the 8th band which contains the widest spectrum of visible light (it contains about two third of the spectrum) has 15 meters resolution (each pixel shows 225 $m^2$ on the ground), so it has a good resolution as seen in Table 1. This band named panchromatic and is considered as the highest resolution band in the OLI sensor [3].

As follows, we overview some fusion approaches, however due to the concentration on another process (*i.e.* interpolation [4] [5] [6]), this overview will be short. Optical/passive remote sensing satellites provide multispectral (MS) and panchromatic (PAN) images which have different spatial, spectral, radiometric, temporal resolution/accuracy. The multispectral images have high spectral information and low spatial information, whereas the PAN image has lower spectral and high spatial information. Fusion of the low (spatial) resolution multispectral images and the high (spatial) resolution PAN images has been a hot problem. The synthetic high spatial and spectral resolution MS image can provide increased interpretation capabilities and more reliable results [7] [8]. During the last two decades, various fusion approaches have been developed. These methods are mainly divided into four categories: projection-substitution-based methods, relative spectral contribution methods, ARSIS-based fusion methods and model-based methods [8] [9]. In [9] differences among these schemes have explicitly been explained. The rest of this paper is organized as follow. Second section expresses the foundations of the proposed method, third section overviews a specific interpolation algorithm entitled linear minimum mean square error-estimation (LMMSE), the forth section evaluates the proposed idea for better image fusion in the Landsat-8 sensor and final section is conclusion of all topics discussed in this paper.

**Table 1.** Spectrum of OLI.

| Band No. | Spectral Band | Wavelength Range (micro meter) | Spatial Resolution (meter) |
|---|---|---|---|
| 2 | Blue (B) | 0.450 - 0.515 | 30 |
| 3 | Green (G) | 0.525 - 0.600 | 30 |
| 4 | Red (R) | 0.630 - 0.680 | 30 |
| 8 | Panchromatic (PAN) | 0.500 - 0.680 | 15 |

## 2. Fundamentals and an Insight on the Proposed Method

There are two principle approaches in colorization of grayscale images. First approach contains the ways which use virtual color to produce color images. Still the only way of colorization in some specific applications is to use these techniques although they are mainly ancient. The second approach contains the ways that use another color image to produce color in a grayscale image. Generally, this color image might be produced by other tools or the same tools. In this article, the second way will be studied which means that we want to colorize the grayscale image with a color image which has been provided by the same sensor. This process is named MS image fusion or pan-sharpening and previously with emphasis on details of fusion technique has been performed, but we are going to apply the concept of interpolation process that did not have suitable attentions in the past studies. In fact, we see many important software tools such as ENVI and ERDAS as the most famous remote sensing image processing tools have only classical interpolation techniques (such as BL and CC) and therefore ENVI- and ERDAS-based researches in image fusion area and even other fusion researches often don't use new and better interpolators. So we only focus on interpolation impact in fusion quality in a specific application, *i.e.* Landsat multispectral images. Obviously, these images don't need to the geometrical corrections because they have been provided by the same sensor. The important point in these images is fusing high and middle resolution images. We use an elementary-based way called IHS [10] for the color production process in colorless images. **Figure 1** shows general processing steps in image fusion and also the IHS fusion technique used here. We have shown the importance of interpolation in this figure as an internal process. According to the IHS technique, we transform the colored image from RGB to HSI color space. Then color information which is in hue and saturation attributions, are taken and after resizing with the panchromatic image, they are fused with this image and the result converts to RGB space again. Color transformations have different ways in various color spaces. Due to the use of HSI (HSV) color space model in this article [11] [12],

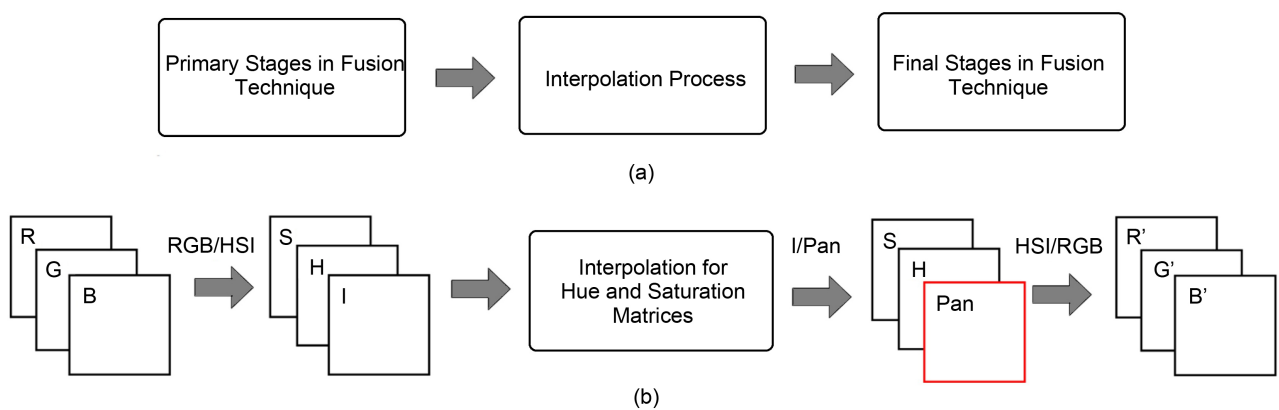

(a)

(b)

**Figure 1.** IHS technique; part (a) shows the general process in all fusion techniques and part (b) represents the processing steps in the IHS Technique.

the Kruse and Raines method [13], which is optimal for USGS images, is used for IHS processing.

Therefore in addition to the IHS method, another question should be answered which is the resizing issue [14]. This is possible with use of specific interpolation algorithm based on a statistical linear estimation. The property of this method in comparison to the traditional interpolation technique like CC method is to consider this point that Landsat images can be considered as the images with MRF-based behavior in the most geographical regions. Totally, all the interpolation methods often work based on a type of statistical averaging and this action is similar to a low pass filter (LPF), so better methods must be searched in the estimation of missing data. The methods must use effective information for estimation of all the correlated pixels (with an appropriate impact coefficient, see **Figure 2(a)**).

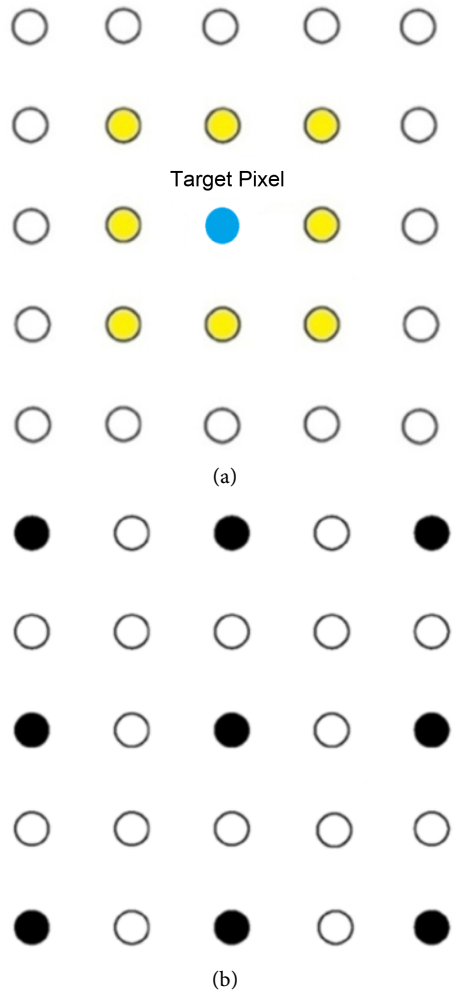

**Figure 2.** Part (a) shows eight pixels with the most impact on the target pixel. Part (b) shows magnifying order for resizing of the color components to 4-time bigger than the first size, this operation generates 75% new pixels in the magnified image that must be estimated based on the original pixels (25%) and an interpolator [2].

## 3. LMMSE Interpolator

In this section, we discuss about LMMSE interpolation method which was presented in [15], this scheme is a statistical edge-guided scheme and uses four to six nearest neighbors to estimate pixels and so it has good suitability with MRF-based images, as seen in **Figure 2(a)**. An important point about LMMSE is the adaptive approach in LMMSE in comparison to the classical schemes such as cubic convolution (CC) technique which uses sixteen nearest neighbors for pixel estimation. And bi-linear (BL) that by use of only four nearest neighbors estimates the target pixel, therefore it is a simple scheme and suitable for the MRF-based images in terms of computational complexity and the impact of neighboring pixels, however it is not adaptive because the experimental results based on grey-scale and color images show that BL is mainly weaker than CC in practice. Outputs of [15] prove LMMSE in grayscale images is better than CC and we will also show in IHS fusion technique (as a color space process), LMMSE is better than CC again. In this study, we should build four times bigger images from two color attributions' images which have 30 meters resolution (namely, each of their dimensions is two times smaller than its equal grayscale image). To achieve this, we have to estimate 75% of the pixels with using 25% of them. There are several interpolation methods for this estimation but in this specific application with considering the use of affine linear transform for resizing/re-sampling. Therefore, we can use a regular interpolation based on statistical estimation which is derived from LMMSE (in some texts entitled DFDF [15] [16] [17]). **Figure 2(b)** shows the transmission of image's pixels in smaller sizes to a 4-time bigger space. We discuss about LMMSE statistical estimation in continuation of this part. As seen in **Figure 3(b),** to calculate the directional value of non-existing pixels, firstly in two angles 45 and 135 degrees, two average amounts for sample pixel in location $x_h(2i, 2j)$ and the same positions will be calculated which are named $x_{45}$ and $x_{135}$ and are derived from Equation (1). Now, we can introduce two error values for these two directions and use them in the next computations. Equation (2) shows the desired error values. In Equation (2), $x_h(2i, 2j)$ or in brief $x_h$ doesn't exist, but it can be obtained towards this point that estimation error becomes at least. We name the $x_h$ estimation $x'_h$, and this estimation can be achieved in this way that the errors will be minimized. Therefore based on the LMMSE method, to achieve such results, according to the Equation (3), $x'_h$ can be known as a linear combination of weights $w_{45}$ and $w_{135}$ of $x_{45}$ and $x_{135}$ where the estimation error reaches the least amount with choosing the appropriate weights $w_{45}$ and $w_{135}$.

$$x_{45} = \frac{x(i, j+1) + x(i+1, j)}{2}$$
$$x_{135} = \frac{x(i, j) + x(i+1, j+1)}{2} \quad (1)$$

$$e_{45}(2i, 2j) = x_{45}(2i, 2j) - x_h(2i, 2j)$$
$$e_{135}(2i, 2j) = x_{135}(2i, 2j) - x_h(2i, 2j) \quad (2)$$

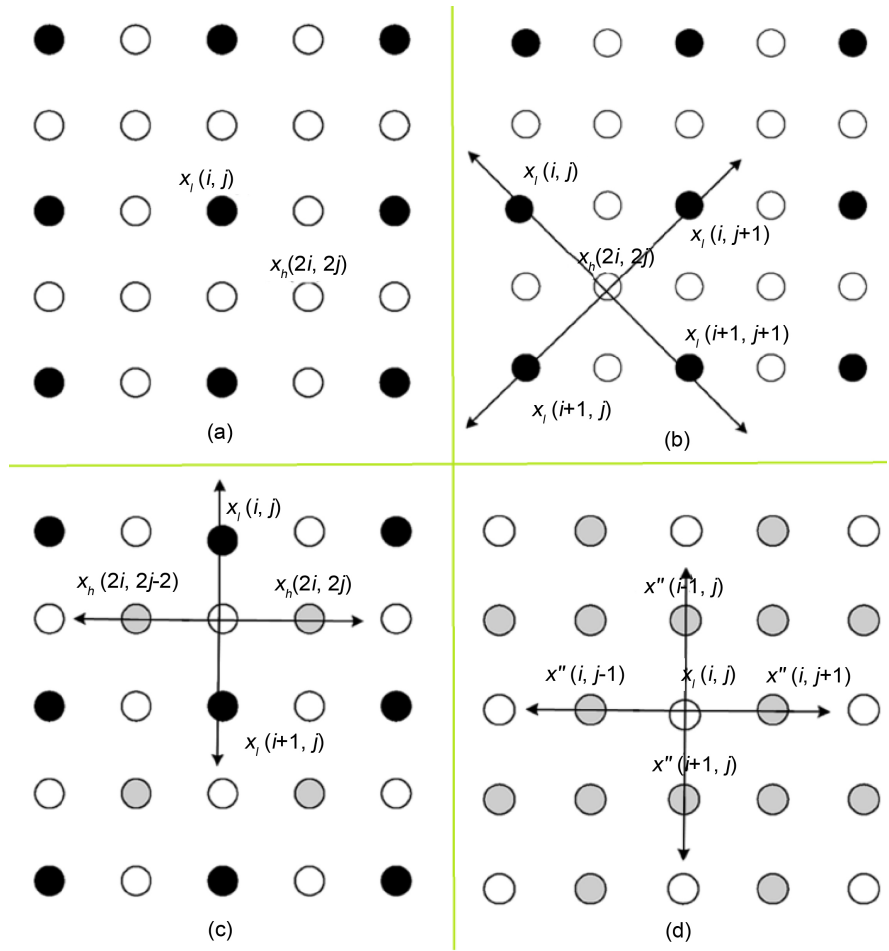

**Figure 3.** This figure shows the steps of estimation of the non-existing pixels (parts (a) to (d), respectively) [2].

Weighs will be as Equation (4) after the necessary calculations. After the calculating $w_{45}$ and $w_{135}$, directional error variances are calculated and the variance calculations will be according to Equation (5) where $u$, $S_{45}$ and $S_{135}$ are achieved from Equations (6) and (7), respectively. For other pixels which are in other positions, from two orthogonal directions (0 and 90 degrees) which the pixels amounts were available from before or achieved as the results of statistical estimation in **Figure 3(b)**, we estimate the non-existing pixels. The continuation of estimation of non-existing pixels will be continued to complete interpolation of all of new pixels (75%) that are shown in **Figure 3(c)** and **Figure 3(d)**.

$$x_h' = w_{45}x_{45} + w_{135}x_{135}$$
$$w_{45} + w_{135} = 1 \tag{3}$$
$$\left\{ w_{45}, w_{135} \right\} = \underset{w_{45}+w_{135}=1}{\arg\min}\, E\left\{ \left( x_h' - x_h \right)^2 \right\}$$

$$w_{45} = \frac{\sigma^2\left( e_{135} \right)}{\sigma^2\left( e_{45} \right) + \sigma^2\left( e_{135} \right)}$$
$$w_{135} = \frac{\sigma^2\left( e_{45} \right)}{\sigma^2\left( e_{45} \right) + \sigma^2\left( e_{135} \right)} = 1 - w_{45} \tag{4}$$

$$\sigma^2(e_{45}) = \frac{1}{3}\sum_{k=1}^{3}(S_{45}(k) - u)^2$$

$$\sigma^2(e_{135}) = \frac{1}{3}\sum_{k=1}^{3}(S_{135}(k) - u)^2$$
(5)

$$u = \frac{x_{45} + x_{135}}{2}$$
(6)

$$S_{45} = \{x(i, j+1), x'_{45}, x(i+1, j)\}$$

$$S_{135} = \{x(i, j), x'_{135}, x(i+1, j+1)\}$$
(7)

## 4. Simulations

To evaluate the proposed method, two panchromatic images shown in **Figure 4(a)** and **Figure 4(b)**, with 15 meters resolution, are colorized with color images from combination of 2nd, 3rd, and 4th bands of the OLI, with 30 meters resolution, shown in **Figure 4(c)** and **Figure 4(d)**. The colorized outputs (fused images) are shown in **Figure 4(e)** and **Figure 4(f)**, respectively. The noticeable quality of the proposed method is based on using the appropriate interpolation. Although in fusing MS images by use of IHS technique, correction for the color weights is often done because of the effect of IR band in panchromatic, see [10] for more information, but it is not used here. For precise assessment of images quality, we should use some appropriate metrics. We know that a visual assessment is not enough; therefore, it is necessary to apply the metrical (computational) quality assessment (QA) along the visual quality assessment. Metrical quality assessment which has two types objective QA and subjective QA, is itself an independent research field and lots of researchers have proposed many different metrics for various applications in image QA (IQA). The used quality assessment metric in here named SSIM, by Wang *et al.* [18]. This metric is completely based on human visual system (HVS) and natural sense statistics (NSS) [19]. The SSIM which was presented by Wang is also a full reference metric (for example the PNSR as a most common quality assessment metric for similarity evaluation is a full reference metric) which means the original image has to be accessible. And in addition, it doesn't have a good matching with HVS and therefore this application is not appropriate because the reference image is not completely accessible. This problem was solved by Yang *et al.* [20] and it is possible to use this metric in such applications. **Figure 5** shows the collections of output image pixels for comparison to the main color image, this process is named down-sampling and is a common way for assessment in digital image processing. Totally, the metrical quality assessment has three types: full reference, reduced reference and no reference [18] [21]. The second state is discussed in this article. SSIM has been shown in Equation (8). In this equation, amounts and $\sigma_x$ and $\sigma_y$ are standard deviations for them, respectively, and also $\sigma_{xy}$ is small non-negative amounts, see more information in this respect in [18] [22]. The SSIM amount is at least equal to −1 or 0% similarity, and at most +1. In SSIM, maximum similarity is corresponding to the 1 or 100% similarity (uniformly). Table 2 shows the SSIM outputs for two fused images of Figure 4(e) and Figure 4(f). The quality of the fusing based on the LMMSE interpolator is obviously confirmed by SSIM. This quality just shows the matching in original image colors and fused ones. The color quality of LMMSE-based fusion is shown in Table 2. However, a pre-interpolation for LMMSE has been used like [15] for improving the performance, see [15] in this regard (pre-interpolation may not be a real-time process). Figure 6 shows the results of the proposed method in comparison to cubic convolution (CC) [23] as a classical method [24].

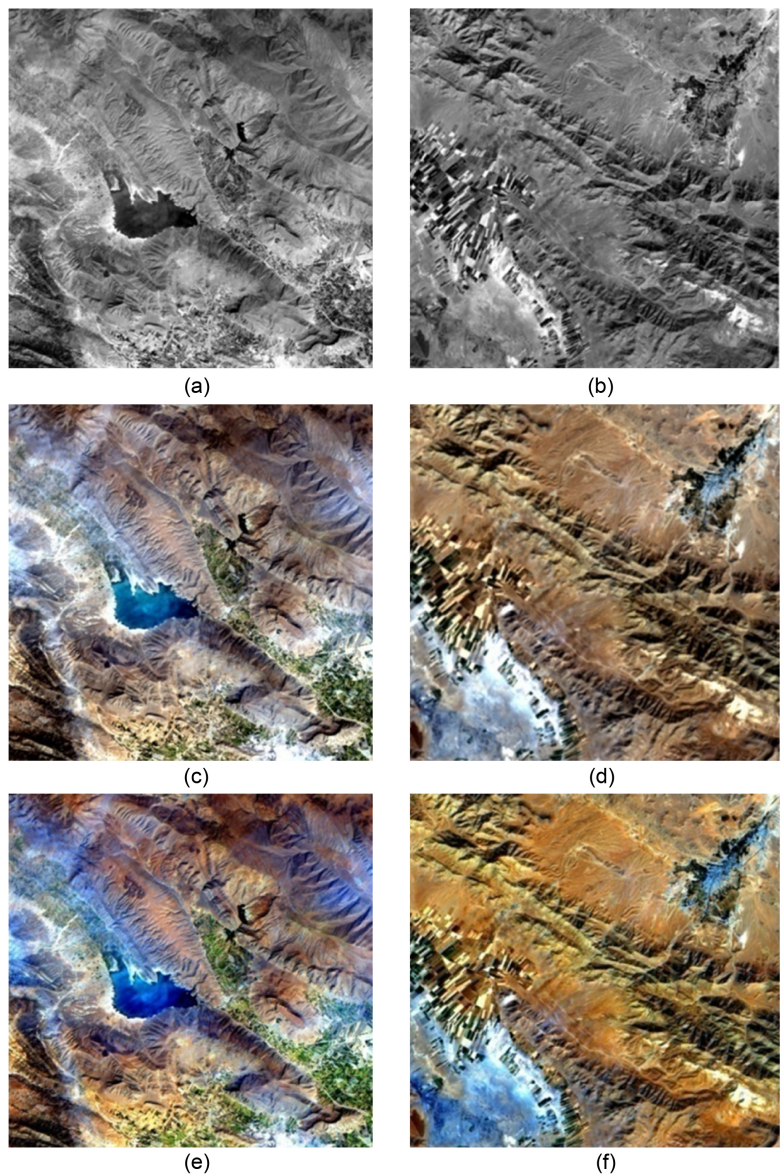

(a)          (b)

(c)          (d)

(e)          (f)

**Figure 4.** Lake (parts (a), (c) and (e)) and Mountain (parts (b), (d) and (f)).

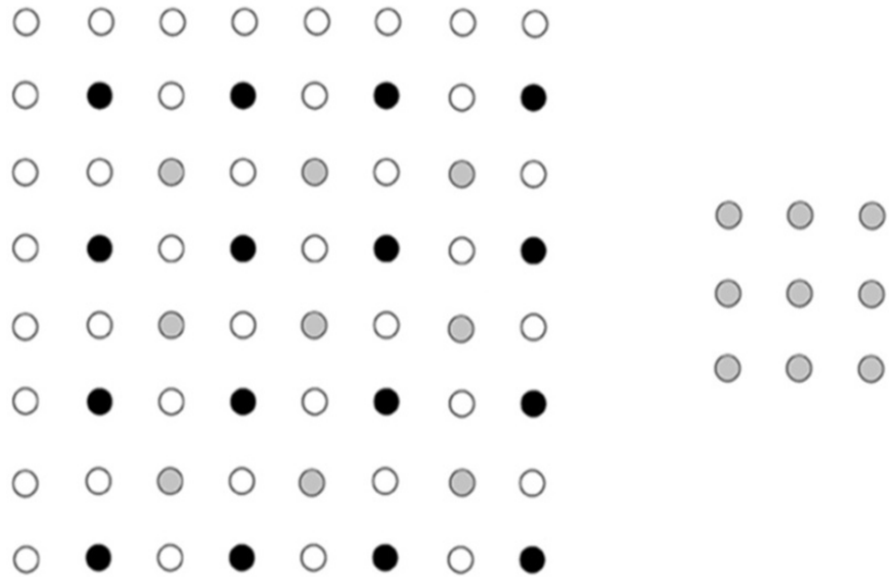

**Figure 5.** Down-sampling for testing the fused images; brown samples are selected for QA.

$$SSIM = \frac{\left(2u_x u_y + C_1\right)\left(2\sigma_x \sigma_y + C_2\right)\left(\sigma_{xy} + C_3\right)}{\left(u_x^2 + u_y^2 + C_1\right)\left(\sigma_x^2 + \sigma_y^2 + C_2\right)\left(\sigma_x \sigma_y + C_3\right)} \tag{8}$$

**Table 2.** Quality assessment in the fused images using SSIM.

| Color Band | Lake | Mountain |
|---|---|---|
| Red | 0.7010 | 0.5713 |
| Green | 0.7165 | 0.5862 |
| Blue | 0.6669 | 0.5691 |
| Average | 0.6948 | 0.5755 |
| Similarity (%) | 84.74 | 78.77 |

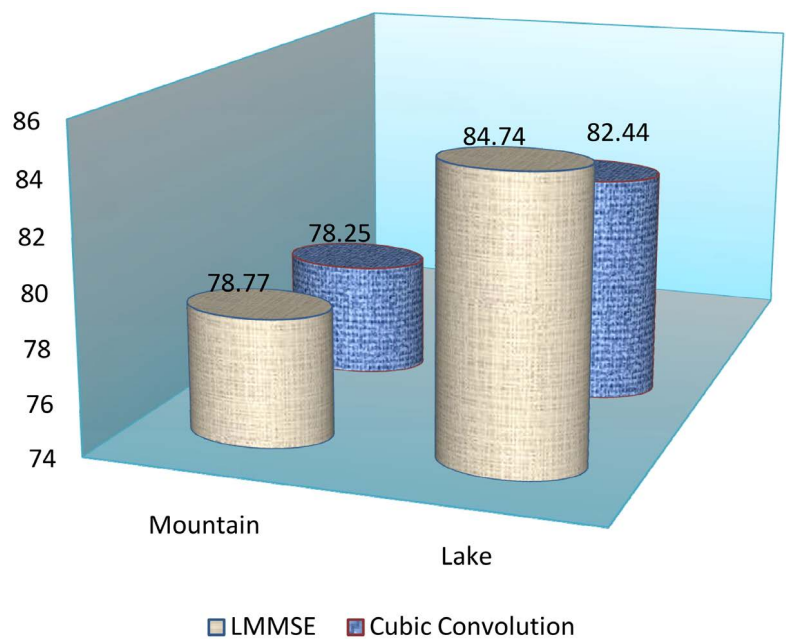

**Figure 6.** Performance for two interpolation methods used in the simulations (with similarity percentage).

## 5. Discussion

One of the best ways for enhancing complex issues is innovative heuristic solutions. Almost proving these innovative solutions are not mathematically possible, except with the simulation, namely, the problems are solved by using simulation-based solutions [25]. In this study, we see the IHS fusion like a color space process guarantees the better quality of the LMMSE interpolator than the CC. According to hard accessibility to high resolution images, it is necessary to use the easy to achieve images in the best way. The discussed subject in this article is the fusion of grayscale images with color samples with using a statistical interpolator. The mentioned method according to the existence of certain relevance between panchromatic image dimensions and MS optical image dimensions of other sensors is easily extensible to them.